# Deep Neural Networks based Meta-Learning for Network Intrusion Detection

Anabia Sohail[1,2], Bibi Ayisha[1], Irfan Hammed[1], Muhammad Mohsin Zafar[1,3], Hani Alquhayz*[4] and Asifullah Khan*[1,5]

[1]Pattern Recognition Lab, Department of Computer & Information Sciences, Pakistan Institute of Engineering & Applied Sciences, Nilore, Islamabad 45650, Pakistan
[2]Department of Computer Science, Faculty of Computing and Artificial Intelligence, Air University, E-9, Islamabad, Pakistan
[3]Faculty of Computer Science and Engineering, Ghulam Ishaq Khan Institute of Engineering Sciences and Technology, Topi 23640, District Swabi, Khyber Pakhtunkhwa, Pakistan
[4]Department of Computer Science and Information, College of Science in Zulfi, Majmaah University, Al-Majmaah,11952, Saudi Arabia
[5]PIEAS Artificial Intelligence Center (PAIC), Pakistan Institute of Engineering & Applied Sciences, Nilore, Islamabad 45650, Pakistan

Corresponding authors: Asifullah Khan (e-mail: asif@pieas.edu.pk), Hani Alquhayz (e-mail: csz@mu.edu.sa)

**ABSTRACT** The digitization of different components of industry and inter-connectivity among indigenous networks have increased the risk of network attacks. Designing an intrusion detection system to ensure security of the industrial ecosystem is difficult as network traffic encompasses various attack types, including new and evolving ones with minor changes. The data used to construct a predictive model for computer networks has a skewed class distribution and limited representation of attack types, which differ from real network traffic. These limitations result in dataset shift, negatively impacting the machine learning models' predictive abilities and reducing the detection rate against novel attacks. To address the challenges, we propose a novel deep neural network based Meta-Learning framework; INformation FUsion and Stacking Ensemble (INFUSE) for network intrusion detection. First, a hybrid feature space is created by integrating decision and feature spaces. Five different classifiers are utilized to generate a pool of decision spaces. The feature space is then enriched through a deep sparse autoencoder that learns the semantic relationships between attacks. Finally, the deep Meta-Learner acts as an ensemble combiner to analyze the hybrid feature space and make a final decision. Our evaluation on stringent benchmark datasets and comparison to existing techniques showed the effectiveness of INFUSE with an F-Score of 0.91, Accuracy of 91.6%, and Recall of 0.94 on the Test+ dataset, and an F-Score of 0.91, Accuracy of 85.6%, and Recall of 0.87 on the stringent Test-21 dataset. These promising results indicate the strong generalization capability and the potential to detect network attacks.

**INDEX TERMS** Autoencoder, Deep Meta-Learner, Deep Neural Networks, Deep Stacking Ensemble, Intrusion Detection, Information Fusion

## I. INTRODUCTION

The emergence of industry 4.0 and beyond, has brought a massive paradigm shift by digitizing essential components of industry and bringing automation. It is grounded on the integration of different technologies and network interconnectivity. The network heterogeneity in industry increases the challenge of security [1]. Rapid adaptation of the robotic automation systems, smart control centres, and distributed smart grids etc. by the leading industries is now more than ever. Moreover, cutting edge research and deployment of high bandwidth network infrastructure such as 45/5G/6G has enabled the general industry as well as software industry to move towards more of a service(s) based and data centric approach such as Internet of Things, and smart devices. All of this has significantly increased the size of global networks. Data privacy and secure communication is a key concern of the industrial ecosystem and indigenous users now more than ever.

Moreover, the wave of COVID-19 pandemic caused a great paradigm shift in internet based activities due to transformation of education, trade, businesses and industrial-setup to online mode. The rapid transition makes network communication an important aspect of human life in different dimensions [2], [3]. Due to this transition the global network experienced a rapid growth of network users from 56.7% to 62.5%. The internet globally connects a vast number of communication devices, physical infrastructure and social networks. The operational control systems, monitoring and controlling the development of industrial machinery as well as the large social networks globally are all interconnected. The network attacks on these systems can result in the loss of the critical data, the assets of strategic importance, or the network control systems culminating into a danger to the national industrial sector [4].

The continuous expansion of the global network confronted with increased risk of sophisticated attacks and network vulnerabilities [5]–[8]. Network attacks and data breaches lead to theft of privileged information and disruption of vital services that can cause financial loss. In 2015, Ukraine faced a power breakdown due to a cyber-attack on the electrical power network. Recently, during the Russia-Ukraine war, the digital infrastructure of Ukraine was crippled by cyber-attacks [5].

The large population relying on the global network makes network security a primary area of focus, as a considerable amount of privacy-sensitive data is being generated and distributed across multiple network nodes. Network traffic is targeted by a diverse range of attacks, including Probing, Denial of Service, User to Root, SQL injections, Cross-site Scripting, Web attacks, and many others [9]. In November 2021, a DDoS attack three times larger than previous records was initiated by approximately 15,000 bots, with a peak throughput of 3.45 Tbps. The security of network traffic is ensured by deploying Intrusion Detection Systems (IDS). However, designing an intelligent IDS for provision of security in industry is challenging due to the distributed nature of networks and diverse types of attacks.

The growing need for network security and the continuous evolution of intrusion methods demands active research in the development of NIDS. Machine learning (ML) techniques are seen as a valuable tool for developing anomaly-based NIDS because of their ability to model anomalous behaviour [10]–[12]. However, the learning paradigm of ML techniques assumes that the probability distribution of the test set is the same as the training set [13]. In real-world applications, especially in network traffic, the data distribution often shifts between training and testing data. The continuous emergence of new and variant attack types poses a challenge of dataset shift in intrusion detection systems [14]. This raises concerns about poor performance of ML models on the test set due to underlying inductive bias of the model.

Dataset shift has received relatively little attention in the context of IDS, despite the fact that new attacks are emerging every day. Therefore, there is a pressing need to develop a technique that can effectively address the non-stationary nature of attacks. It is challenging to choose a model with the right bias and optimal hypothesis space that can accurately generalize the deviation in attack distribution. Thus, it is necessary to develop a technique that can model the deviation pattern of attacks, rather than focusing solely on specific attack types. Meta-learning is one of the useful techniques that is based on learning to learn and improves the decision by distilling knowledge from the learning experience of multiple learners [15].

This work proposes a novel deep Metal-Learning based Information Fusion and Stacking Ensemble framework (INFUSE) by performing both decision- and feature-level information fusion to address the dataset shift problem in network intrusion detection. The performance of the proposed INFUSE is assessed on the benchmarked network traffic dataset. The distinct contributions of this work are:

- *Diverse feature spaces were explored to model the data distribution, which can address unseen emerging attack variations. The strong representational learning ability of weight regularized deep sparse autoencoders is utilized to encode the semantic relevance of normal and malicious traffic.*
- *A strong decision space was produced by utilizing multiple base-classifiers with varying hypothesis spaces to reduce the inductive bias of each hypothesis space.*
- *A deep neural network based Meta-Learner is used that systematically evaluates a multi-information carrying hybrid feature space to differentiate between attacked and normal samples.*
- *The proposed INFUSE has undergone evaluation by comparing its performance with several existing techniques and different ensembles. The results of the performance assessment indicate its strong generalization ability, which is not limited to specific attack types.*

The paper proceeds by Background literature on intrusion detection in Section 2. The dataset is described in Section 3, and methodology of the proposed ensemble is presented in Section 4. The results are analyzed in detail in Section 5 and finally, the significance of the proposed technique is concluded in Section 6.

## II. BACKGROUND LITERATURE

In the past, several techniques have been developed for the analysis of network traffic, which involve either binary classification of intrusions in network traffic or multiclass segregation of attacks into specific categories. To achieve this, different approaches have been explored, including classical ML, deep learning, and ensemble learning techniques, utilizing publicly available datasets [16]–[18]. A brief overview of the existing techniques is provided below.

Initially, classical ML techniques were used to detect attacked samples in network traffic by focusing on the extraction or selection of informative representations from the data. Aslahi-Shahri et al. [19] developed a hybrid model that used genetic algorithms to select an optimal feature subset from the NSL-KDD dataset, followed by SVM classification. The technique achieved a 0.97 F-score by selecting 10 features out of 45. In another study, Liu et al. [20] used both unsupervised and supervised learning to address network intrusion. They first identified similar groups within the dataset using k-means clustering, and then used a Random Forest (RF) algorithm to classify these clusters as normal or attacked traffic. They followed this up with a multi-class classification of attacked samples using deep CNN and LSTM. This approach achieved accuracies of 85% and 99% for NSL-KDD and CIS-IDS2017 datasets, respectively. In another study [21], feature embeddings are generated from network packets and assigned to SVM for better classification. The multiverse optimizer is used for Benmessahel et al. for parameter optimization of neural network as a binary classifier [22]. One major limitation of classical ML techniques is their reliance on different feature extraction techniques to deal with dataset variance effectively. A highly efficient intrusion detection system requires immediate detection of any attack.

In recent times, deep neural networks have emerged as a formidable tool for network traffic analysis and a variety of captivating applications in image processing. Their unparalleled capacity for representation mapping has garnered significant attention among researchers. Many researchers have utilized deep neural networks for network traffic analysis due to their strong representation mapping capacity [23], [24]. These systems often follow two steps. In the first step, feature representation is learned using deep neural network and classified in the second step. Qureshi et al. [25] exploited the idea of deep autoencoders for binary classification of network traffic. In this study, initially new feature representation was extracted from deep sparse autoencoder based on self-taught learning and an augmented representation was generated by concatenating with original feature space. Finally, an encoder of an autoencoder was trained for classification. Al-Qataf et al. [26] similar to previous approaches used autoencoder for extracting low-dimensional embedding from network traffic. An autoencoder was pretrained by exploiting self-taught learning for generation of embedding which was assigned to SVM for classification.

In recent times, deep neural networks have emerged as a formidable tool for network traffic analysis and a variety of captivating applications in image processing. Their unparalleled capacity for representation mapping has garnered significant attention among researchers [27]–[30]. In another study, the performance was improved by combining four different CNN architectures [31]. The feature space of the NSL-KDD dataset was transformed into an image to take advantage of the learning capacity of CNN. Four different CNN architectures were merged before the fully connected layer and trained with a single loss function. The final decision was based on a 256-dimensional feature space, but the use of multiple CNNs increased the complexity. Naseer et al. [32] investigated various techniques, including deep CNN, LSTM, and AE, to differentiate between attacked samples and normal data flow. However, due to the significant difference between the attack samples in the training and testing sets, these techniques exhibit a low detection rate for attacked samples.

Ensemble learning has shown promise in improving individual classifiers for handling noise sensitivity, domain shift, scalability, and inability to detect diverse attacks. Gao et al. developed an adaptive learning-based ensemble to overcome the complexity of intrusion datasets [33]. Five different classifiers, including decision tree, RF, kNN, and DNN, were used as base learners in the ensemble. Majority voting was used to make a decision by assigning a weightage to each classifier's decision. This ensemble technique was used for intrusion detection in NSL-KDD Test+. Similarly, another study suggested an ensemble of n modified Adaboost algorithms. Cost-sensitive base-classifiers were developed by optimizing the Adaboost algorithm using the area under the curve to mitigate the class imbalance challenge. Salo et al. selected significant features by applying PCA and Information Gain [34]. They enhanced the learning capacity by developing an ensemble of SVM, instance-based learning algorithms, and multilayer perceptron.

Zhang et al. [35] proposed using multiple feature fusion and homogenous stacking ensemble mechanisms to detect irregularities in network traffic. A diverse set of features was generated to train n number of homogenous base classifiers. The predictions of the base classifiers were combined using RF as a meta-classifier to draw a final decision. The idea of heterogeneous ensembles has been proposed to enhance the learning capacity by addressing the shortcomings of homogenous ensembles. Zhou et al. developed a voting-based heterogeneous ensemble. Initially, they used a hierarchical feature extraction algorithm CFS-BA to boost the feature representation at the preprocessing step. The proposed approach exploited shallow algorithms such as Forest Penalizing Attributes, C4.5, and RF on the extracted representation, and an average voting strategy was used to combine the base classifiers' probabilities [36].

The main issue with the existing approach is that it ignores the non-stationary nature of attacks and uses accuracy as the performance metric, which can underestimate the detection rate of the minority class. In real network traffic, a wide range of attacks exist, and the proportion of attacks is often imbalanced with respect to normal traffic. Therefore, it is essential to take into account the issue of dataset shift and prioritize the detection ability of the ML model for highly imbalanced malicious attack data.

TABLE I.
Feature representation of NSL KDD dataset

| Category | Description | Datatype |
|---|---|---|
| *Basic Features* | Features representing TCP/ICP connection without considering the payload | Symbolic & Continuous |
| *Content Features* | Features required to access the payload of TCP packet and suspicious behaviour within payload | Symbolic & Continuous |
| *Time based Traffic Features* | Network features were analyzed with 2s temporal window and provide a statistical information | Continuous |
| *Host based Traffic features* | Feature used to analyze the attack within interval longer than 2s | Continuous |

## III. BACKGROUND LITERATURE

NSL-KDD dataset was utilized in this study for the analysis of network traffic intrusion. It is considered a benchmark dataset for the development of NIDS and appropriate for estimation of model's performance for dataset shift problem [37]. NSL-KDD dataset consists of features extracted from the header and payload of the network traffic packets. The dataset's records are classified as normal or attacked, with the attacked samples falling into five main categories commonly found in network traffic. Table I provides a description of the dataset's characteristics. The following characteristics make it suitable for analyzing the effectiveness of NIDS against a dataset shift problem:

- *There is no repetition of records in the train and test set contrary to other datasets. Sample recurrence in the test set makes the performance evaluation criterion unrealistic. NSL-KDD dataset makes an equal contribution to each sample to remove bias.*
- *The train and test sets are highly imbalanced that increases the difficulty level. The frequency of normal and attacked traffic in the Train, Test+ and Test-21 are represented in Fig. 1.*
- *The difficulty level was assigned to both the train and test set. The additional stringent dataset Test-21 was generated by the test examples that were misclassified by the 21 classifiers.*
- *Test set was stringent and has samples of attacks that are not included in training set. The training set includes normal data and 22 attack types, while the test datasets (Test+, Test-21) contain seven new attack types in addition to those present in the training data. The presence of unique attacks in the test set motivates the development of a model that can handle emerging attacks and the challenge of distribution shift.*

### A. Statistical Evaluation of Dataset Shift

In this work, the Kolmogorov-Smirnov test is applied to analyze a dataset shift between train and test set. It is a non-parametric test and compares how likely that two datasets are drawn with the same probability distribution [38]. Hypothesis test for comparison of test distribution with train is formulated as:

$H_0$: *train and test sets are sampled from same data distributions*

$H_A$: *train and test sets have different data distributions*

The Kolmogorov-Smirnov test computes test statistics D by regressing the empirical distribution function of the test set on cumulative distribution function of reference dataset (train data) using the following formula:

$$D_{n,m} = max\,|F(x) - E(x)| \quad (1)$$

In (1), n is the sample size of cumulative distribution function $(F(x))$ of the train set, whereas m is the sample size of empirical distribution function $(E(x))$ of the test set. The test statistics D = 0.0128 show that there is a significant difference between two distributions with significant p-value (p-value= 6.82e-280 < 0.05).

### B. Feature space based Evaluation of Dataset Shift

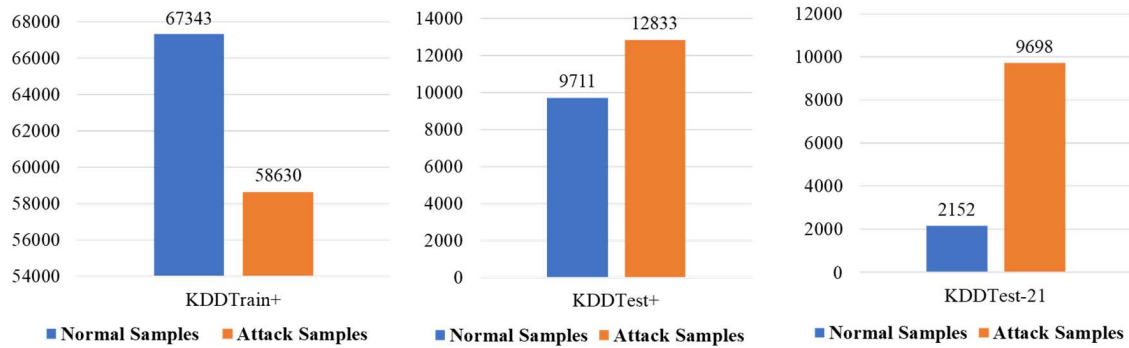

FIGURE 1. The proportion of attacked samples in the train and test sets (Test+ and Test-21).

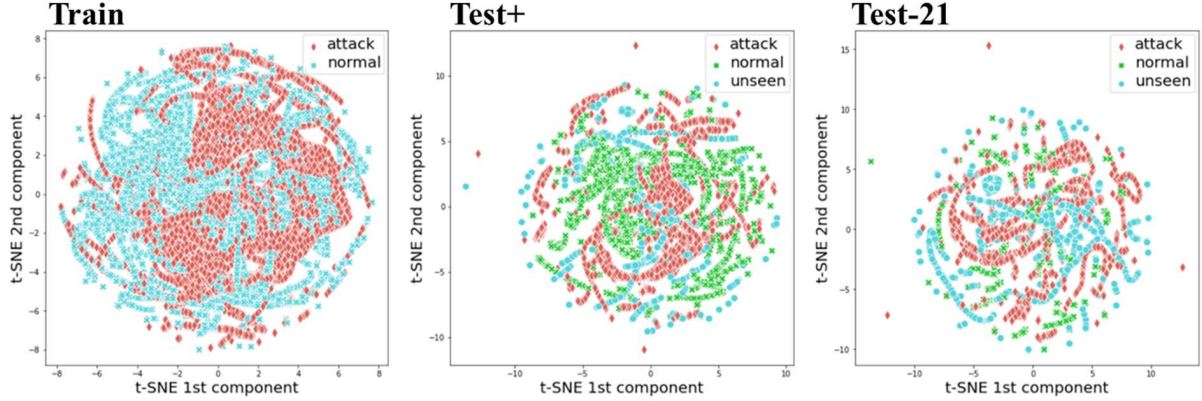

FIGURE 2. Visualization of the first two components of the feature space projected via t-SNE.

Feature space of the train $p_{train}(X_{tr})$ and test set $p_{test}(X_{ts})$ is visualized using t-SNE (Fig. 2) to qualitatively analyse the shift in distribution. Fig. 2 shows that the distribution of the test set is not only different from the train set $p_{train}(x_{tr}|y_{tr}) \neq p_{test}(x_{ts}|y_{ts})$ but also that the test set has some unique attacks $p_{train}(y_{tr}|x_{tr}) \neq p_{test}(y_{ts}|x_{ts})$ that are not part of the training set.

### C. Dataset Representation and Preprocessing
NSL-KDD dataset comprises 41 network features that are measured on continuous and ordinal scales. The symbolic features are represented by indicator variables in this study, which increases the feature dimension from 41 to 121.000000.
The NSL-KDD dataset has a highly skewed feature distribution, leading to biased variable contributions and reduced classifier performance. To address this issue, we applied data normalization to scale the Train, Test+, and Test-21 sets while preserving their original distributions.

### D. Dataset Division
In this study, we used a stratified hold-out strategy to randomly split the training dataset into a 60:40% for training of base-classifiers and the Meta-Learner, respectively. We select an optimal set of hyperparameters for base-classifiers by performing 5-fold cross-validation on the 60% reserved training data. The Meta-Learner is trained on 80% of the 40% split, while 20% is utilized for validation and hyperparameter selection. Finally, we assess the generalization performance of the proposed ensemble on separately provided test sets (test+ and test-21) by NSL KDD.

## IV. THE PROPOSED DEEP META-LEARNING BASED *IN*FORMATION *FU*SION AND *S*TACKING HETEROGENEOUS *E*NSEMBLE FOR NID

The network data is heterogeneous in nature and vulnerable to various types of attacks, posing a challenge to develop a robust NIDS that can handle dataset shift problems. The learning scenario of NIDS is specified as the development of a classification model $F_\theta(\cdot)$ that inputs network traffic sample as $x_i$ and learns a mapping function $h_\theta(\cdot)$ that makes a binary classification between normal $N_i$ and attack $T_i$ samples. The network traffic consisted of normal and attack samples corresponding to N and $A_1$, $A_2$, $A_3$, $A_4$, $A_n$, respectively. Different types of attacks are represented as $A_1$, $A_2$, $A_3$, $A_4$, and $A_n$. In general ML scenario, the distribution of training and test set is same, i.e. $p_{train}(x_{tr}|y_{tr}) = p_{test}(x_{ts}|y_{ts})$. However, when there is shift between train and test set distribution ($p_{train}(x_{tr}|y_{tr}) \neq p_{test}(x_{ts}|y_{ts})$) then ML model will result in poor generalization due to underlying conditions. In the case of dataset shift, the learning scenario of NIDS is different from conventional approach as distribution of training samples such as N, $A_1$, $A_2$, and $A_3$ are considered by ML is different from test set that violates the learning condition of supervised ML model (shown in Fig. 3).

To address this deficiency, we proposed a deep NN based Meta-Learning framework named *IN*formation *FU*sion and *S*tacking *E*nsemble (INFUSE). The proposed framework is formulated as a model $F_M()$ that instead of focusing on a specific attack type generalizes the intrusion behaviour from training samples (N, $A_1$, $A_2$, $A_3$). The proposed framework works sequentially in two phases: first, developing an information-rich feature space, and second, developing a Meta-Learner to generate a final decision function from multiple information. The concept of information fusion is utilized at both the feature and decision level to tackle the dataset shift problem. Fig. 4 provides a detailed workflow of the proposed deep Meta-Learning based framework INFUSE.

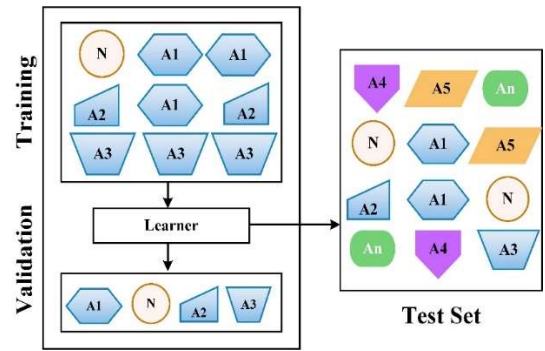

FIGURE 3. Dataset shift scenario.

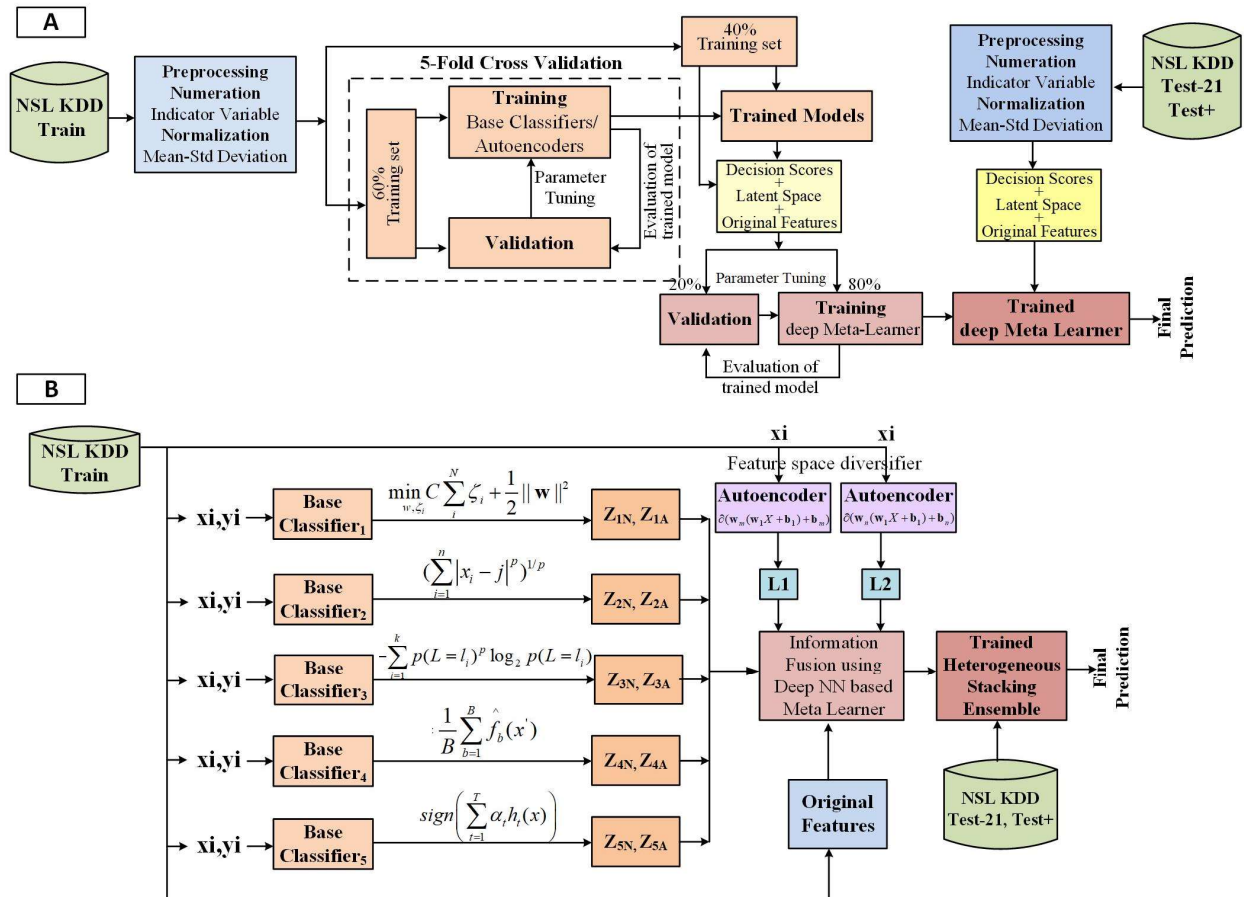

FIGURE 4. Panel (a) shows training and testing phases, whereas panel (b) depicts detailed layout of the proposed INFUSE.

### A. DECISION SPACE DIVERSIFICATION MODULE

The idea of stacking heterogeneous ensemble is utilized to enhance the generalization of the NIDS. We have used five heterogeneous base classifiers $H = \{H_1, H_2, H_3, H_4, H_5\}$ and combined their decision Score $Z = \{Z_{1N}, Z_{1A}, Z_{2N}, Z_{2A}, Z_{3N}, Z_{3A}, Z_{4N}, Z_{4A}, Z_{5N}, Z_{5A}\}$ to generate a pool of diverse hypothesis varying in their learning biases. A machine learning model has an inherent bias towards a particular data distribution, which can affect its performance when dealing with dataset shift. The pool of decision spaces will help in overcoming the bias and variance associated with individual learners [39]. Thus, the instances that cannot be tackled via single learner can be corrected by other classifiers [40], [41]. Diversity in the hypothesis space of base classifiers is discussed below. The parameters of the base classifiers are mentioned in Table II.

TABLE II
Parameter Setting of Base Classifiers

| Base Classifier | Parameter Values |
|---|---|
| Base classifier-1 | Classifier: SVM, Kernel: Rbf, C: 100, Gamma: 0.01 |
| Base classifier-2 | Classifier: kNN, Metric:Eucledian, Distance: weight, $k$:3 |
| Base classifier-3 | Classifier: Decision Tree, Criterion: entropy, Features: 94 |
| Base classifier-4 | Classifier: Random Forest, Criterion: entropy, Features: 12, Estimators: 39 |
| Base classifier-5 | Classifier: Adaboost, Estimators: 12 |

#### 1) BASE-CLASSIFIERS' DECISION POOL

The first base classifier was selected based on the property of structural risk minimization. In this regard, SVM is used [42] that draws an optimal hyperplane by incorporating geometric representation of the hyper-plane during training, mathematically represented in (2) & (3). The second base classifier kNN was nonparametric in nature and target function was locally estimated instead of explicit modelling of target function based on the whole training space. This approach considers class label of $k$ nearest neighbours for assigning a label to new attack as shown in (4) thus, shows promising results for classification of unknown and unique attacks [43]. Rule-based approach is used in the third base classifier to appropriately deal with categorical and nominal datasets [33]. For this, the decision tree is implemented using entropy and it is mathematically represented in (5). Furthermore, an idea of ensemble learning is employed in base classifier-4 using Random Forest (RF). RF improves robustness towards attacks by manipulating both data distribution and features during training. RF draws the final decision by taking an average of all the predictions made by the forest of trees [44]. It is mathematically expressed in (6). AdaBoost is used as a base classifier-5 that reduces the bias of a classifier [5] by performing the serial training of a set of $n$ weak base-learners and weighted sampling during training. Thus, each classifier focuses on correctly classifying the samples misclassified in the previous iteration.

$$w^T x + b = 0 \qquad (2)$$

$$\min_{w,\zeta} C \sum_{i}^{N} \zeta_i + \frac{1}{2}\|w\|^2 \qquad (3)$$

$$D(X,j) = (\sum_{i=1}^{n} |x_i - j|^p)^{1/p} \qquad (4)$$

$$H(L) = -\sum_{i=1}^{k} p(L = l_i)^p \log_2 p(L = l_i) \qquad (5)$$

$$\hat{R} = \frac{1}{B}\sum_{b=1}^{B} \hat{f}_b(x') \qquad (6)$$

In (2), an input instance is represented by $x$, whereas $w^T$ is a weight vector and $b$ are a bias. In (3) misclassifications made by a SVM are represented by $\zeta_i$, and $C$ is a hyper-parameter that establishes tradeoff between generalization and empirical error on a training set. Eq. (4) shows a distance function for kNN that compares a test instance with a training instance using Euclidean distance. In (5), $p(L = l_i)$ is the probability of an element for a given class $i$ for decision tree. In (6), $B$ is the number of decision trees that are considered during prediction, and $b$ is the current decision tree. Whereas data sub-set is defined by $x'$, $f_b(.)$ is the function that fits the decision tree on feature subset, and final output is denoted by $R$.

### B. FEATURE SPACE DIVERSIFICATION MODULE
In this work, we improved the feature space by training two AEs in an unsupervised manner to learn semantics of the data and generated a new representation reflecting the underlying distribution of network traffic. In this regard, a new attack instance that lies outside the distribution can be modelled based on semantic relevance. The newly generated latent representation $L = \{L_1, L_2\}$ is combined with the original feature space $X$ to provide more discriminative features $F = f_c(L_1 \parallel L_2 \parallel X)$.

#### 1) WEIGHT REGULARIZED DEEP SPARSE AE
We exploited two weight regularized deep sparse AEs with depth of 8- and 10- layers to capture the latent representation of attacks. The description of the proposed AEs is mentioned in Table III and Fig. 5. AE is an effective representational learning algorithm based on unsupervised NNs. The learning principle of AE is based on identity function and follows the encoder-decoder paradigm [45]. The mathematical representation of AE is shown in (7) & (8).

$$\hat{L}_1 = f_{enc}(X) = \partial(w_n(w_{n-1}..(w_1 X + b_1) + b_{n-1}) + b_n) \qquad (7)$$

$$X' = f_{dec}(\hat{L}) = \partial(w'_n(w'_{n-1}..(w'_1 \hat{L} + b'_1) + b'_{n-1}) + b'_n) \qquad (8)$$

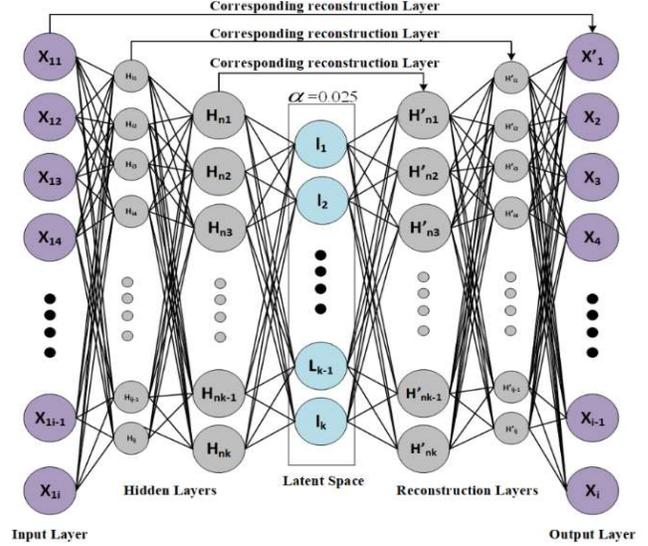

FIGURE 5. Generalized schematics of the proposed weight regularized AE.

In (7) $X$ is the original input, $W = \{w_1,...,w_n\}$ represents weights, $B = \{b_1,...,b_n\}$ is the added biases. AE maps $X$ to a latent representation $L$ using an encoding function $f_{enc}(.)$. In (8), the reconstructed input is represented by $X'$ that is generated from the latent representation ($L$) using the decoding function $f_{dec}(.)$.

In this work, we enhance the representation learning ability of the AE by making it deep and adding the sparsity penalty. Deep AE maps the original input to useful representation by hierarchically learning multi-level feature representations while each hierarchy level corresponds to a different level of abstraction. The L2 weight regularization was applied as a sparsity penalty in the mean square error loss function that enforces the AE to generate the effective and compact representation from the network traffic packets.

$$MSE = f_{error}(\frac{1}{n}\sum_{i=1}^{n}\|X_i - X'_i\|^2 + \lambda\|w\|^2) \qquad (9)$$

The sparsity penalty for AE is shown in (9). L2 weight regularization penalizes the model in proportion to the magnitude of the weights. Thus, it enforces weights near zero and makes AE to learn a reduced feature space by discovering the structure hidden in the high dimensional feature space. A latent representation is generated by learning a semantic relevance between features corresponding to the same group.

Table III.
Parameter setting of the proposed weight regularized autoencoder and Meta-Leraner.

| Model | Layers | Learning rate | Optimizer | Loss | Weight decay |
|---|---|---|---|---|---|
| INFUSE | 6 | 0.00008 | SGD | BCELoss | - |
| AE-1 | 10 | 0.00008 | Adam | MSELoss | 1e-5 |
| AE-2 | 8 | 0.0001 | Adam | MSELoss | 1e-5 |

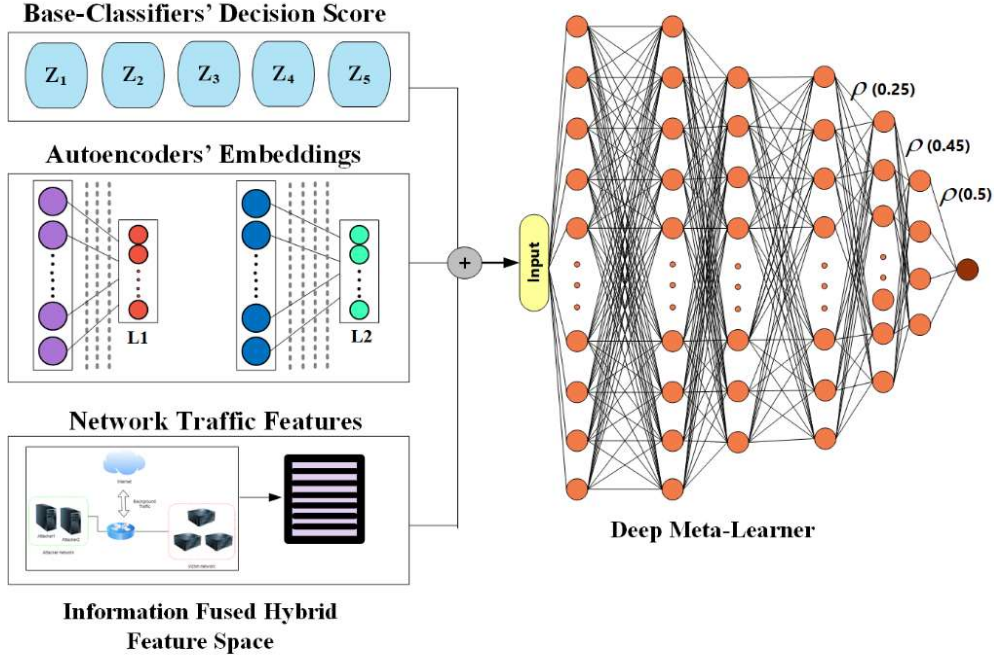

FIGURE 6. Generalized schematics of the proposed deep NN based Meta-Learning Framework.

## C. THE PROPOSED DEEP META LEARNER

A deep Meta-Learning framework is used to learn a discrimination function $F_M(\cdot)$ for binary classification of attacked samples. A 6-layers deep, fully connected NN is used as a Meta-Learner (Fig. 6) that performs the information infusion by integrating diverse feature representations with multiple learning experiences and analysing the information-rich feature-space collectively in an intelligent way for final decision. The proposed Meta-Learner is deep in architecture that makes use of multiple levels of nonlinearities to generate a strong representative context from information rich feature space.

## D. THE PROPOSED DEEP META LEARNER

The parameter setting of the Base and Meta-Learner is performed on the validation dataset that is separated from the training set. Tables II and III show the parameter on which proposed models give the optimal performance.

## V. RESULTS AND DISCUSSION

The significance of the proposed ensemble INFUSE against dataset shift has been analyzed on the Test+ and Test-21 of the NSL-KDD dataset. The performance has been evaluated using several performance measures and compared with other techniques.

### A. EVALUATION MEASURES

Multiple evaluation metrics, including accuracy (Acc.), F-score, detection rate (Recall), false negative rate (FNR), ROC, and PR curves, are considered to estimate the performance of the proposed ensemble INFUSE against dataset shift. Detection rate is used to evaluate the model's capacity to recognize attacks, and Accuracy and F-score are also reported. The statistical significance of the proposed ensemble is assessed via the McNemar test, and the standard error is computed using z-statistics and confidence interval was set to 95%. True Positives ($T_P$) are the samples correctly recognized as attacks, and True Negatives ($T_N$) are the number of normal instances correctly classified. Accuracy

TABLE IV
Performance evaluation of the proposed ensemble INFUSE and baseline classifiers.

| Technique | NSL-KDD Test+ | | | NSL-KDD Test-21 | | |
|---|---|---|---|---|---|---|
| | F-Score±error | Acc. (%) | Recall | F-Score±error | Acc. (%) | Recall |
| *Proposed INFUSE* | 0.91±0.003 | 91.64 | 0.94 | 0.91±0.005 | 85.6 | 0.87 |
| *MLP* | 0.84±0.005 | 83.05 | 0.77 | 0.79±0.007 | 69.18 | 0.69 |
| *Deep NN* | 0.87±0.004 | 84.70 | 0.89 | 0.86±0.006 | 76.90 | 0.84 |

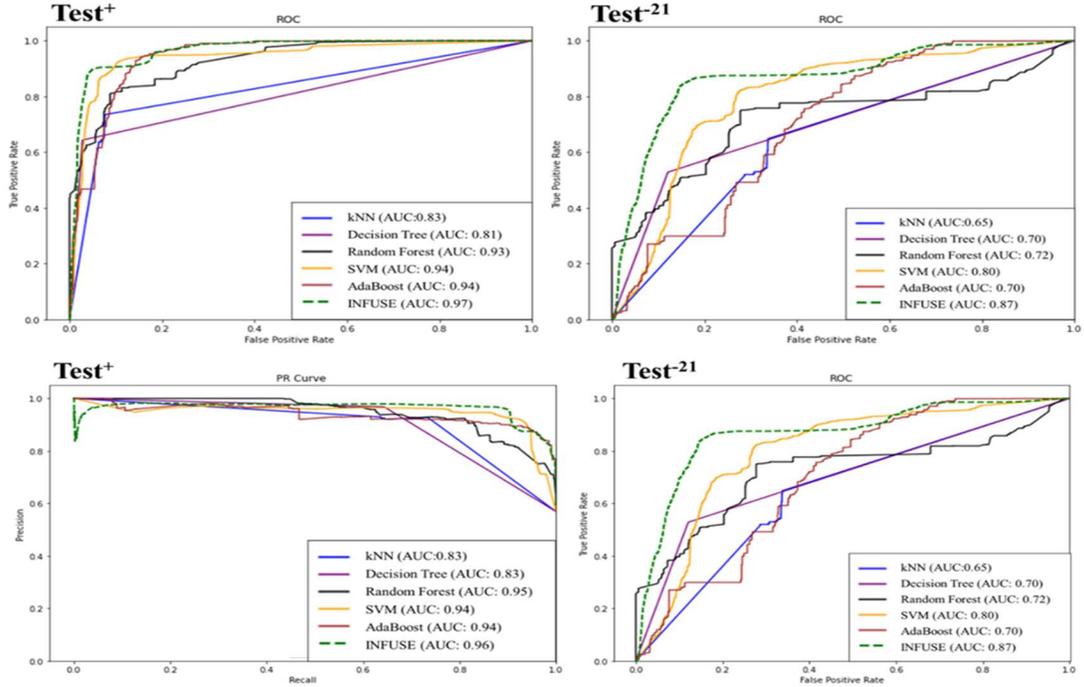
FIGURE 7. Graphical illustration of performance using ROC and PR curves.

defines the percentage of correct predictions regardless of the normal and attacked class. The performance measures are defined as follows.

$$DetectionRate(Recall) = \frac{T_p}{T_p + F_N} \quad (10)$$

$$Specificity = \frac{T_N}{T_N + F_P} \quad (11)$$

$$Accuracy = \frac{T_P + T_N}{T_P + T_N + F_P + F_N} \quad (12)$$

$$F - Score = 2 \cdot \frac{Precision \times Recall}{Precision + Recall} \quad (13)$$

$$FNR = \frac{FN}{TP + FN} \quad (14)$$

**B. PERFORMANCE ESTIMATION OF INFUSE**

Initially, the performance of the proposed INFUSE was analyzed by comparing it to standard baseline models, namely MLP (2-layers deep) and a deep fully connected neural network (6-layers deep). Table IV shows the F-score, accuracy, and recall, indicating that the proposed ensemble INFUSE outperforms the baseline models on both datasets. The remarkable performance of INFUSE on the Test-21 dataset, which has a high distribution shift, suggests that the proposed technique is effective in addressing the dataset shift problem.

**C. PERFORMANCE COMPARISON OF INFUSE WITH BASE CLASSIFIER**

The performance of the base-classifiers is presented in Table 5, where SVM has the highest accuracy of 82.80% and 67% on Test+ and Test-21, respectively. The F-scores of base-classifiers range from 0.69 to 0.83 on Test+ and 0.55 to 0.76 on Test-21. While the base-classifiers show reasonable performance on KDD Test+, their performance deteriorates on KDD Test-21, a stringent dataset with a high class imbalance of unseen attacks that are difficult to classify. The empirical analysis in Table V suggests that the proposed ensemble INFUSE outperforms the base-classifiers and significantly addresses the problem of dataset shift.

TABLE V
Performance evaluation of the base classifiers and the proposed ensemble INFUSE.

| Technique | NSL-KDD Test+ | | | NSL-KDD Test-21 | | |
|---|---|---|---|---|---|---|
| | F-Score±error | Acc. (%) | Recall | F-Score±error | Acc. (%) | Recall |
| *Proposed INFUSE* | 0.91±0.003 | 91.64 | 0.94 | 0.91±0.005 | 85.6 | 0.87 |
| *Base classifier-1 (SVM)* | 0.83±0.005 | 82.80 | 0.73 | 0.76±0.0076 | 67.41 | 0.64 |
| *Base classifier-2 (KNN)* | 0.76±0.0055 | 76.76 | 0.64 | 0.66±0.0085 | 55.83 | 0.53 |
| *Base classifier-3 (Decision Tree)* | 0.77±0.0055 | 78.48 | 0.64 | 0.68±0.0083 | 59.16 | 0.53 |
| *Base classifier-4 (Random Forest)* | 0.69±0.006 | 72.65 | 0.53 | 0.55±0.009 | 48.30 | 0.39 |
| *Base classifier-5 (AdaBoost)* | 0.76±0.0055 | 76.56 | 0.64 | 0.66±0.0085 | 55.49 | 0.53 |

TABLE VI
Ablation study of the proposed ensemble learning technique on KDD Test+.

| Technique | NSL-KDD Test+ | | | | |
|---|---|---|---|---|---|
| | F-score±error | Acc. (%) | Recall | Specificity | FNR |
| Proposed decision scores based ensemble | 0.83±0.005 | 82 | 0.755 | 0.92 | 0.245 |
| Proposed feature stacked ensemble (Decision scores + Original features) | 0.89±0.004 | 89 | 0.88 | 0.89 | 0.115 |
| Proposed feature stacked ensemble (Original + AE features) | 0.86±0.0045 | 84 | 0.82 | 0.87 | 0.177 |
| Proposed feature stacked ensemble (Decision scores + Original + 1 AE features) | 0.91±0.0037 | 90.30 | 0.93 | 0.87 | 0.074 |
| Proposed INFUSE | 0.91±0.003 | 91.64 | 0.94 | 0.90 | 0.063 |

TABLE VII
Ablation study of the proposed ensemble learning technique on KDD Test-21.

| Technique | NSL-KDD Test-21 | | | | |
|---|---|---|---|---|---|
| | F-Score±error | Acc. (%) | Recall | Specificity | FNR |
| Proposed decision scores based Ensemble | 0.77±0.0075 | 67.10 | 0.78 | 0.58 | 0.21 |
| Proposed stacked ensemble (Decision Scores + Original Features) | 0.871±0.006 | 79.50 | 0.85 | 0.56 | 0.152 |
| Proposed feature stacked ensemble (Original + AE features) | 0.83±0.0067 | 74.58 | 0.77 | 0.61 | 0.223 |
| Proposed feature stacked ensemble (Decision Scores + Original + 1 AE features) | 0.898±0.005 | 83.28 | 0.90 | 0.52 | 0.098 |
| Proposed INFUSE | 0.91±0.005 | 85.6 | 0.87 | 0.78 | 0.126 |

*D. STATISTICAL ANALYSIS*

McNemar's test is applied to statistically evaluate the significance of the proposed ensemble INFUSE [46]. The hypothesis test is formulated as:
$H_0$: the proposed ensemble is equivalent to base-classifier in performance.
$H_1$: the proposed ensemble performs better than base-classifier.
The performance is accessed by comparing performance with the best performing base-classifier using Eq. (15).

$$\chi^2 = \frac{(b-c)^2}{b+c} \quad (15)$$

In the above equation, $b$ is the number of misclassifications made by the base-classifier whereas $c$ is the total misclassification by ensemble model.
The test statistics suggest a statistically significant difference in performance of the proposed ensemble compared to best performing base-classifier with statistics = 566.7, p-value = 2.88e-125 < 0.005 for test+ and statistics = 753.5, p-value = 6.803e-166 < 0.005 for test-21.

*E. ROC AND PR CURVE BASED ANALYSIS*
The ROC curve graphically illustrates the ratio of the true positive rate to the false-positive rate at multiple confidence intervals, while the PR curve is for the imbalanced dataset that shows the ratio of correctly classified examples among the positively predicted samples at several confidence intervals. As shown in Fig. 7, the performance of the proposed ensemble improves on unseen data. Figure 7 also provides a graphical illustration of AUC-ROC and AUC-PR on Test+ and Test-21.

*F. ABLATION STUDY*
We also conducted a thorough analysis to assess the importance of in formation fusion in the proposed ensemble by excluding each of the feature spaces one by one. The results are presented in Tables VI and VII, which indicate that the proposed INFUSE shows better performance on both the datasets than that of the individual feature spaces. The results show that the proposed technique considerably gains an increase in F-score and accuracy on Test-21 and achieves a good trade-off between recall and specificity.

*G. ANALYSIS OF THE SIGNIFICANCE OF THE PROPOSED INFUSE*

TABLE VIII
Performance comparison of the proposed ensemble with different ensemble learning strategies on Test+.

| Technique | Type | NSL-KDD Test+ F-Score±error | NSL-KDD Test+ Acc. (%) | NSL-KDD Test-21 F-Score±error | NSL-KDD Test-21 Acc. (%) |
|---|---|---|---|---|---|
| *Proposed INFUSE* | Stacking Ensemble | *0.91±0.003* | *91.64* | *0.91±0.005* | *85.6* |
| *Proposed Hybrid space with SVM Meta-Learner* | Stacking Ensemble | 0.78±0.0055 | 78.80 | 0.69±0.0083 | 59.76 |
| *Proposed Hybrid space with XGBoost Meta-Learner* | Stacking Ensemble | 0.76±0.0055 | 77.69 | 0.66±0.0085 | 57.59 |
| *Max-weighted voting* | Classical Ensemble | 0.79±0.0053 | 80.00 | 0.72±0.0080 | 62.05 |
| *Average voting* | Classical Ensemble | 0.768±0.0055 | 78.00 | 0.67±0.0084 | 58.40 |
| *Majority voting* | Classical Ensemble | 0.77±0.0055 | 78.38 | 0.67±0.0084 | 58.89 |

The robustness of our proposed ensemble INFUSE against dataset shift challenge was estimated by assigning the information-fused hybrid feature space to SVM and XGBoost as Meta-Learners. Additionally, a performance comparison was made with different classical ensemble techniques, and the results are shown in Table VIII. The empirical evaluation demonstrated that the proposed deep learning-based Meta-Learner significantly outperformed other Meta-Learners as well as max-weighted, average, and majority voting-based ensemble techniques.

### H. DETECTION RATE AND FEATURE SPACE VISUALIZATION

In IDS, detecting each type of attack is crucial for system security. Fig. 8 shows the detection rate and FNR for the proposed ensemble and other techniques. As the type and nature of attacks change every day, IDS should be robust against new attack types. Therefore, the detection rate of the proposed ensemble INFUSE is analyzed for all the unique attack profiles included in the test sets. In the test set, there are 7 attack profiles - mscan, processtable, snmpguess, sainl, apache, httptunnel, and mailbomb - that are unique attack variants and were not seen by the classifier during training. The results, shown in Figure 9, propose that the proposed ensemble INFUSE achieve a considerable detection rate for new attacks that were not seen by the classifier during training.

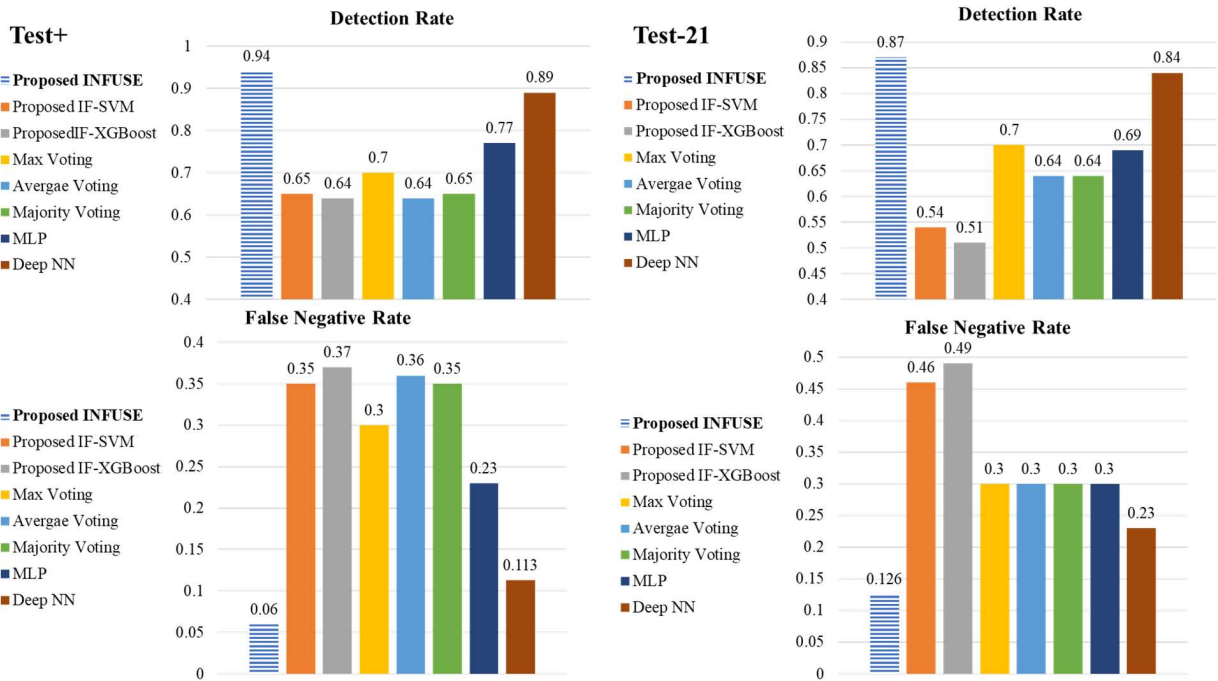

FIGURE 8 Detection Rate and False Negative Rate (FNR) analysis.

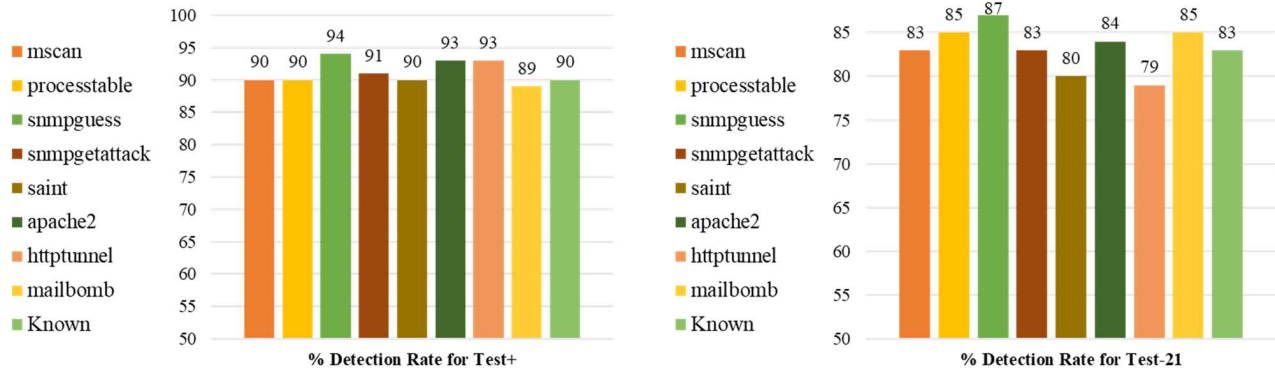

FIGURE 9. Detection Rate analysis of the proposed INFUSE for different attack types.

Feature space visualization using t-SNE shows that the proposed ensemble INFUSE significantly separates the attacked samples from normal traffic samples (Fig. 10).

I. PERFORMANCE COMPARISON WITH EXISITING TECHNIQUES

The proposed ensemble INFUSE has been assessed by making comparison with state-of-the-art studies that employed ML, deep learning, and ensemble learning techniques. The results, shown in Table IX, indicate that the proposed ensemble INFUSE significantly outperforms other classifiers both on Test+ and Test-21.

## VI. CONCLUSION

Network security threats have increased with the rapid expansion of industry 4.0. The security and privacy of data is an important concern of organizations and the industrial ecosystem. Intelligent intrusion detection techniques are required to ensure the network integrity and reliability to maintain the normal workflow of the industrial ecosystem. This study presents a robust stacking heterogeneous ensemble using the idea of information fusion to address the problem of dataset shift and poor generalization in intrusion detection. Our proposed approach, INFUSE, improves the detection rate by utilizing weight regularized deep sparse autoencoders, while achieving specificity towards attacks through exploitation of multiple decision spaces. In the decision stage, a deep neural network-based Meta-Learner intelligently draws a final decision from the hybrid feature space, establishing a considerable trade-off between specificity and detection rate. The results in terms of accuracy (Test+: 91.6%, Test-21: 85.6%) and detection rate (Test+: 0.94, Test-21: 0.87) demonstrate the effectiveness of the proposed INFUSE in network intrusion detection. Comparison with other ensemble learning techniques and recently reported techniques highlights the strong detection ability of INFUSE towards unseen attacks. In the future, this approach may prove useful for Zero-day attack detection and other malware analysis problems, enabling improved generalization.

TABLE IX
Performance comparison with existing state-of-the-art techniques.

| Technique | NSL KDD Test+ Acc. (%) | NSL KDD Test-21 Acc. (%) |
|---|---|---|
| *Proposed INFUSE, Information fusion and deep Meta learner* | *91.6* | *85.6%* |
| *Mushtaq et al.* [47], LSTM+AE | 89.84 | - |
| *Zhang et al.* [35] (Stacking Ensemble) | 84 | - |
| *Sham et al.* [48] (Deep CNN) | 81.00 | - |
| *Zhou et al.* [36] (Voting based Ensemble) | 87.37 | 73.57 |
| *Tama et al.* [49] (Two-Stage Ensemble) | 85.8 | 72.52 |
| *Qureshi et al.* [25] (Autoencoder) | 84.60 | 79.90 |
| *Chohan et al.* [50] (Deep CNN) | 89.41 | 80.36 |
| *Singh et al.* [51] (RNN) | 84.03 | 69.75 |
| *Li et al.* [31] (Multi-CNN fusion) | 86.95 | 76.67 |
| *Gao et al.* [52] (Ensemble) | 84.54 | 71.29 |
| *Naseer et al.* [32] (Deep CNN) | 85.00 | 70.00 |
| *Qatf et al.* [26] (Sparse Autoencoder & SVM) | 84.96 | 79.42 |
| *Ashfaq et al.* [53] (Fuzzy based semi supervised Learning) | 84.12 | 68.82 |
| *Yin et al.* [54] (RNN) | 81.29 | 64.67 |
| *Vinet et al.,* [55] (Clustering + SVM) | 91.3 | - |

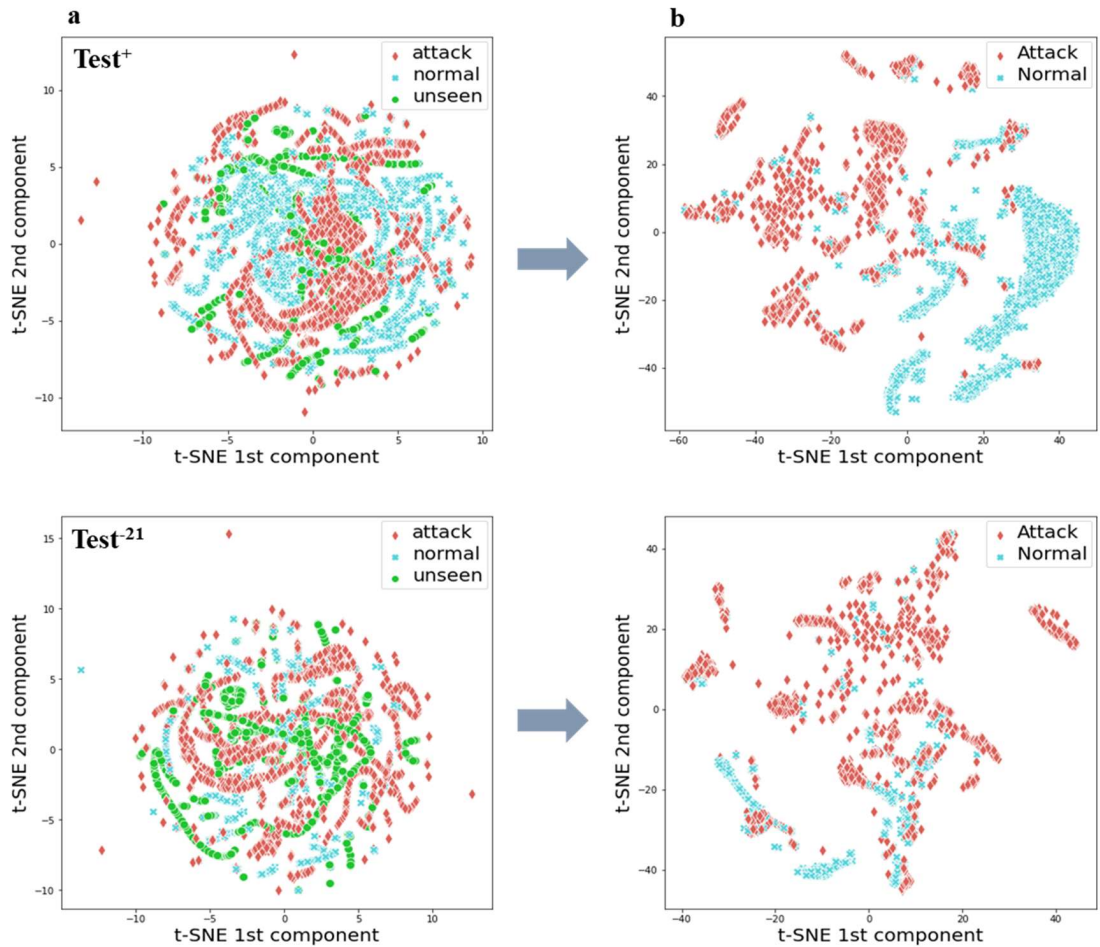

FIGURE 10. Feature space is visualized via t-SNE. Panels (a) and (b) show original and feature space of the last layer of the proposed INFUSE, respectively.

## ACKNOWLEDGMENT
The authors thank Pattern Recognition Lab at Department of Computer and Information Sciences, PIEAS, for providing computational facilities.

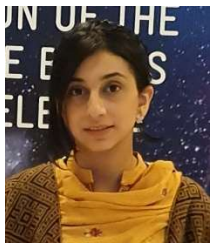
**Anabia Sohail** is working as Assistant professor at department of Computer Science, Air University, Pakistan. She has number of publications in SCI/SCIE indexed journals. She receiver a PhD degree in Computer Science from PIEAS. She received her M.S and B.S degrees in Bioinformatics. Her areas of research are Deep Neural Networks, Ensemble Learning, Biomedical Informatics and data visualization.

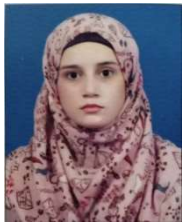
**Bibi Ayisha** has done her MS in Computer Science from NUST, Rawalpindi and BS in Computer Science from Department of Computer & Information Sciences, PIEAS. Her research interests include, Machine Learning, Deep Learning and predictive modeling.

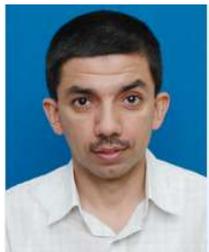
**Muhammad Irfan Hameed** is Incharge of Data Center as well as teaching at Department of Computer and Information Sciences, PIEAS. He has done MS in Computer Software Engineering from NUST, Rawalpindi and BE Electrical Engineering from UET, Taxila. His research Interests are Establishment of Software Services Infrastructure, Development of an Integrated Set of Information Systems, Development of Internet and Database Application Software, Data Warehousing, Information Security Management

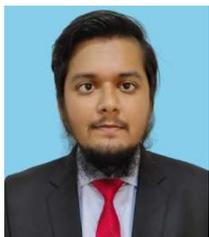
***Muhammad Mohsin Zafar*** received his degree of Bachelor of Engineering in Computer Software Engineering from the National University of Sciences and Technology (NUST), Islamabad, in 2018 and degree of MS in Computer Science from Pakistan Institute of Engineering and Applied Sciences (PIEAS), Islamabad, in 2020. He is currently serving as a faculty member at Ghulam Ishaq Khan Institute of Engineering Sciences and Technology (GIKI). His *research interests* include End to End *Deep Neural Networks, Medical Image Analysis, Multi Loss* Neural Networks, Hybrid Networks, and Machine Learning Model Deployment.

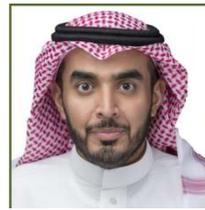
**Hani Alquhayz** received the bachelor's degree in computer science and the master's degree in information systems management from King Saud University, and the Ph.D. degree in computer science from De Montfort University, U.K. He is currently an Associate Professor with the Computer Science Department, College of Science, Majmaah University, Saudi Arabia. He has authored several articles in high-impacted journals, such as IEEE Access, Sensors, and Wireless Communications and Mobile Computing. His research interests include wireless security, scheduling, image processing, the IoT, security and privacy, data mining, machine learning, and artificial intelligence.

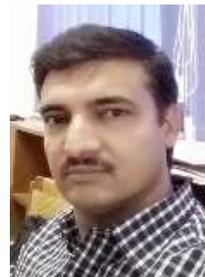
**Asifullah Khan** is a Professor in the Department of Computer and Information Sciences at Pakistan Institute of Engineering and Sciences (PIEAS). He has a distinguished 19-years teaching and research career. Recently, he has received the prestigious "Pride of Performance" award in Computer Science from the President of Pakistan. He received post-doctoral research fellowship from South Korea and carried-out research at Asia's top ranked research institute, GIST. Dr. Khan is currently leading the Pattern Recognition Lab and PIEAS Artificial Intelligence Center. His areas of research are Deep Neural Networks, Machine Learning, Pattern Recognition, Intrusion detection, Medical Image Analysis and Digital Watermarking.